\documentclass{article}
\usepackage{MAIN}
\pdfoutput=1
\usepackage[utf8]{inputenc} 
\usepackage[T1]{fontenc}    
\usepackage{hyperref}       
\usepackage{url}            
\usepackage{booktabs}       
\usepackage{amsfonts}       
\usepackage{nicefrac}       
\usepackage{microtype}      
\usepackage{lipsum}
\usepackage{fancyhdr}       
\usepackage{graphicx} 
\usepackage{caption}
\usepackage{algorithm}
\usepackage{float}
\usepackage{multicol}
\usepackage{amsfonts}
\usepackage{amsmath}
\usepackage{algpseudocode}
\usepackage{amsmath}
\usepackage{tikz}
\usepackage{mathdots}
\usepackage{yhmath}
\usepackage{cancel}
\usepackage{color}
\usepackage{siunitx}
\usepackage{array}
\usepackage{multirow}
\usepackage{amssymb}
\usepackage{tabularx}
\usepackage{extarrows}
\usepackage{booktabs}
\usepackage{changepage}
\usepackage[numbers]{natbib}
\usepackage{graphicx}
\graphicspath{{images/}}     

\DeclareMathOperator{\argmin}{argmin}

\title{RPNR: Robust-Perception Neural Reshading}
\date{April 25, 2022}
\author{
  Fouad Afiouni\thanks{This work was done when the authors were undergraduate \\   students at the American University of Beirut, Lebanon} , Mohamad Fakih\footnotemark[1] \ and  Joey Sleiman\footnotemark[1]
}

\begin{document}

\maketitle

\begin{multicols}{2}

\begin{abstract}
Augmented Reality (AR) applications necessitates methods of inserting needed objects into scenes captured by cameras in a way that is coherent with the surroundings. Common AR applications require the insertion of predefined 3D objects with known properties and shape. This simplifies the problem since it is reduced to extracting an illumination model for the object in that scene by understanding the surrounding light sources. However, it is often not the case that we have information about the properties of an object, especially when we depart from a single source image. Our method renders such source fragments in a coherent way with the target surroundings using only these two images. Our pipeline uses a Deep Image Prior (DIP) network based on a U-Net architecture as the main renderer, alongside robust-feature extracting networks that are used to apply needed losses. Our method does not require any pair-labeled data, and no extensive training on a dataset. We compare our method using qualitative metrics to the baseline methods such as Cut and Paste, Cut And Paste Neural Rendering, and Image Harmonization.
\end{abstract}

\section{Introduction}
\label{sec:intro}

Previous work by Bhattad and Forsyth \cite{cutandpaste} dubbed this problem the "Cut and Paste Neural Rendering" problem since we are essentially cutting an object from a source image and inserting it into another target scene image. This problem is fundamental for image editing and AR applications since such insertions are often needed. However, no automatic or systematic method has been developed yet, which makes the problem require hand-crafting a new shading for the object image, which restricts the results of the hard work to a single object - scene pair.\par

We tackle this problem by developing a framework that can be applied to any such object-scene pair and would output the necessary new shading. However, unlike other popular computer vision problems, this one does not have a high-quality dataset with such corresponding pairs. We thus have to rely on unpaired training of the model using various consistency features to achieve the needed reshading cleanly. These features are often hard to extract in a deterministic setting, which pushes machine learning solutions to the front: we use auxiliary pre-trained models to extract such features and enforce the needed consistencies.

We expand on the work in \cite{cutandpaste} by identifying key areas to improve or otherwise change. Our approach seeks to orthogonalize the training losses that are used to update the DIP, as well as restricting the generation to only the required fields.  
\paragraph{Contributions}
We summarize our novelties as follows:
\begin{itemize}
    \item We base the problem on a new formulation requiring only the shading and gloss fields to be regenerated.
    \item We demonstrate that training the DIP with no external priors achieves good results for this transformation task.
    \item We propose using features robust to illumination for enforcing the consistency losses.
\end{itemize}

\section{Previous Works}
\label{sec:prev}
This section will discuss previous work and studies conducted on the topics of image decomposition, harmonization and relighting. The section will also discuss the related study Cut-And-Paste Neural Rendering by Anand Bhattad and David A. Forsyth on which we build our study.

\textbf{Image decomposition:} The Retinex model, developed by Land, assumes that effective albedo will have large image gradients since it displays sharp and localized changes, while shading has small gradients \cite{Land1, Land2}. The advantage of these models is that they do not require a ground truth. An alternative is image decomposition training using CG rendered images with specialized losses \cite{Li, bi, Fan}. It is also noted that rendering constraints may be used to produce a form of self-supervised training \cite{Janner}. In order to evaluate image decomposition the weighted human disagreement rate is used \cite{Bell}, and as such the current champions are \cite{Fan}. We use an image decomposition method, similar to that of \cite{cutandpaste}, which is built around albedo and shading paradigms to train the image decomposition network without requiring real image ground truth decompositions.

\textbf{Image harmonization (IH)}: IH procedures aim at correcting corrupted images. IH methods are trained to correct images in which an image fragment has been altered by some noise process applied to the original image \cite{Sunk, Tsai, Cong}. These methods could be applied to our study; however, IH methods change the albedo of an inserted object and not their shading since they aim at ensuring the consistency of color representations. This is not the aim of our experiment since we wish to alter the shading.

\textbf{Image Relighting}: Since our renderings are entirely image-based, we cannot use conventional relighting methods that require training data with lighting parameters/environment maps or multi-view data to construct a radiance field \cite{Garon, Hold-Geoffroy,Srinivasan,Mildenhall}. Current single-image relighting methods relight portrait faces under directional lighting \cite{Sun,Zhou,Nestmeyer}. Their approach can relight matte, gloss and specular objects with complex material properties for indoor and outdoor spatially varying illuminated environments, from a single image only, and without requiring physics-based BRDF \cite{Li2}.

\textbf{Cut-And-Paste Neural Rendering}: Cut and Paste methods as described in \cite{cutandpaste} is a method that takes an object from one image and inserts it into another. Cut-and-paste neural rendering is an alternative method to render the inserted fragment’s shading field to become coherent with that of the target scene. In order to train a neural render to render an image with consistent image decomposition inferences, Deep Image Prior (DIP) is used. The results obtained from DIP should have an albedo and shading consistent with cut-and-paste albedo and the target’s shading field. Cut-and-paste surface normals are congruous with the final rendering’s shading field which results in a simple procedure that produces convincing and realistic shading. We build open this method to obtain a procedure that orthogonalizes the training losses that are used to update the DIP, as well as restricting the generation to only the required fields.

\section{Background}
\label{sec:background}
\newcommand{\ftheta}{f_\theta}
\subsection{Deep Image Priors (DIP)}
The work by Ulyanov et al. \cite{DIP} demonstrates that deep network architectures, ConvNets specifically, inherently capture some significant statistics about given images before being trained on any data. They show that ConvNets naturally have high impedance to noise, shuffling, obscuring, and other types of image degradation. Formally, their method dictates the following:\\
Consider a parametrization of the image $x$ with parameters $\theta$, i.e: $x = f_\theta(z)$ where $z$ is a fixed random tensor which is typically less dimensional than $x$. We subsequently refer to this parametrization simply by $f_\theta$. We then try to find optimal parameters $\theta$ to reconstruct the image $x$ after being degraded by some operation that yields $x_0$ \footnote{$x_0$ is a degraded version of the image $x$}. The loss function used to update the gradients of the ConvNet is as follows:

\begin{equation}
    \label{eq:DIP_LOSS}
    \mathcal{L}\left(\theta\right) = \mathbf{E}\left(\ftheta, x_0\right) + \mathcal{R}\left(\ftheta\right)
\end{equation}

The $\mathbf{E}$ term is a standard error term that is most often chosen to be the L2 loss, and the $\mathcal{R}$ term is a regularization term that is often not used in practice \footnote{We refer the inclined reader to the Total Variation Loss, that is often used as the regularizer}. The authors show that updating the parameters $\theta$ to find the function that minimizes $\mathcal{L}$ in \ref{eq:DIP_LOSS}, would result in a function that would yield a good approximation of $x$. In other words:
\begin{align}
    \theta^* &= \argmin_\theta \mathcal{L}\left(\theta\right) \\
    x^* &= f_{\theta^*} \approx x
\end{align}
\subsection{U-Net architecture}

\begin{figure}[H]
    \centering
    \includegraphics[width=\linewidth]{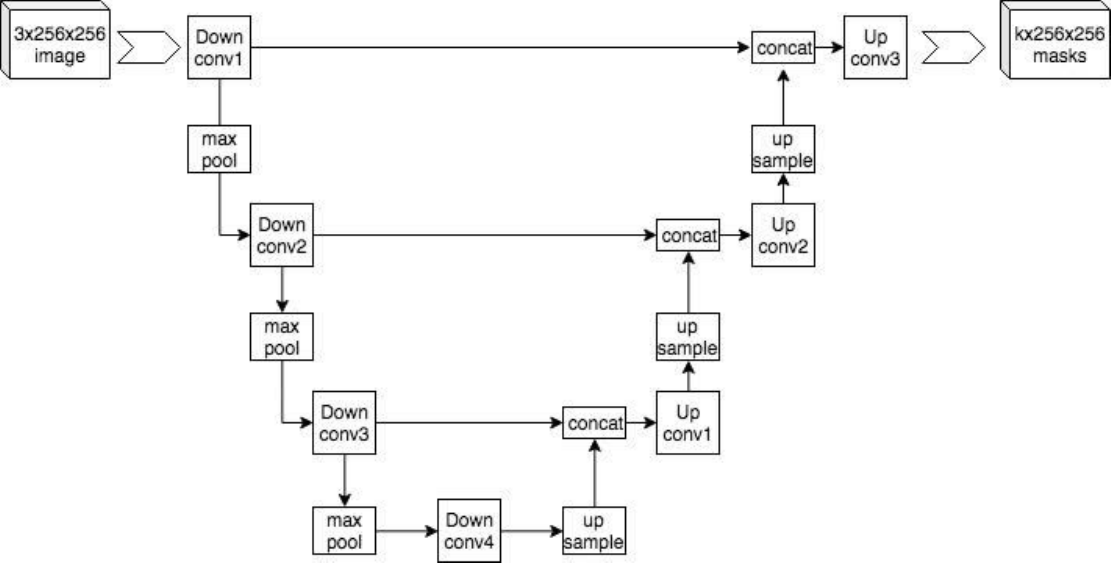}
    \caption{The U-Net architecture}
    \label{fig:unet}
\end{figure}

The U-Net architecture for ConvNets has been the standard go-to method for moving between image domains (pix2pix). In contrast to typical ConvNets where there are consecutive downsampling to extract features, U-Nets are in the family of Encoder-Decoder structures. Early Encoder-Decoder models aimed to extract representative features in a latent space by forcing the data through a bottleneck in the architecture: The initial data is fed as input and the model is trained to yield each input at its output. The down-convolutions form the Encoder, and the up-convolutions form the Decoder. U-Nets expand this definition by including Skip-Connections between corresponding down and up convolutions. A diagram of this is shown in figure \ref{fig:unet}\par

The U-Net architecture has been used extensively in many Computer Vision problems given its flexibility and scalability. It is worth to mention that such an architecture is not limited to direct inference only since it can be used in many settings, such as a GAN generator, or a GAN discriminator.
\subsection{U-Net Discriminators for GANs}GANs were proposed by Goodfellow et Al \cite{GAN} as a  novel generative system that samples from a learned distribution of a given dataset. GANs have been demonstrated to have immense learning power, and have been almost universally dominant over all generative problems and applications. The GAN architecture is composed of two competing models, a Generator and a Discriminator which are placed in an adversarial setting where the generator is trying to approximate the data distribution, and the discriminator is trying to detect whether a sample comes from the true distribution or the learned distribution. This creates a Game Theory problem where eventually the generator (given enough parameters and time) will be able to sample from the same data distribution, and where the discriminator is not able to distinguish between a real and a fake sample. The typical formulation of this problem uses $G(z)$ for the generator where $z$ is simply a prior used, often sampled from a uniform or normal distribution, and $D(x)$ for the discriminator which simply outputs a value of 1 for real, and 0 for fake. Let $\mathcal X$ be the domain of the data distribution, and $\mathcal{Z}$ be the domain of the priors, then the following is a formulation for a two-dimensional data space $\mathcal{X} \subset \mathbb{R}^{n*m}$:
\begin{equation}
        \begin{aligned}
            G &: \mathcal{Z}  \rightarrow \mathbb{R}^{n*m}\\
            D &: \mathbb{R}^{n*m}  \rightarrow \mathbb{R}\\
            D^{*}(x)&=\Bigl\{\ \begin{array}{cc}
                            1 & x\ \in \mathcal{X}\\
                            0 & x=G( z)
                            \end{array}\\
            \mathcal{L} &= \mathbb{E}_{z}[( 1-D( G( z) )] \ +\ \mathbb{E}_{x}[ D( x)]
        \end{aligned}
    \end{equation}
Where $\mathcal{L}$ is the metric function that is updated to solve the following formulation:
\begin{equation}
    \min_G \max_D \mathcal{L}
\end{equation}
The work by Schonfeld et Al. \cite{unetdisc} proposes using a U-Net architecture as a discriminator for Generative Adversarial Networks (GANs). Instead of the generator $G$ only outputing a single scalar representing the "realness" of the entire data sample, the UNet Disciminator $D_u$ outputs a global realness scalar, along with a realness scalar for each data element in the sample. 
\begin{equation}
        \begin{aligned}
            G &: \mathcal{Z}  \rightarrow \mathbb{R}^{n*m}\\
            D &: \mathbb{R}^{n*m}  \rightarrow \left(\mathbb{R}, \mathbb{R}^{n*m}\right)\\
            D\left(x\right) &= \left(D^{enc}, D^{dec}\right)\\
            D^{*}(x)&=\Bigl\{\ \begin{array}{cc}
                            \left(1, \textbf{1}^{n*m}\right) & x\ \in \mathcal{X}\\
                            \left(0, \textbf{0}^{n*m}\right) & x=G( z)
                            \end{array}\\
            \mathcal{L}^{enc} &= \mathbb{E}_{z}[( 1-D^{enc}( G( z) )] \ +\ \mathbb{E}_{x}[ D^{enc}(x)]\\
            \mathcal{L}^{dec}_{ij} &=\mathbb{E}_{z}[( 1-D^{dec}_{ij}( G( z) )] \ +\ \mathbb{E}_{x}[ D^{dec}_{ij}(x)]\\
            \mathcal{L}^{dec} &= \sum_{ij} \mathcal{L}^{dec}_{ij} \\
            \mathcal{L} &= \mathcal{L}^{enc} + \mathcal{L}^{dec}
        \end{aligned}
    \end{equation}
The Game Theory formulation still holds and it is the same optimization problem but we add an extra element per pixel in the output image. The authors also propose using the CutMix augmentation scheme to enforce a desired output consistency (invariance to class-domain transformations).

\section{RPNR Setup}
\subsection{Observations and Assumptions}
    Our designed method aims to circumvent generating required fields if they are extractable from elsewhere, especially if the field in question is needed to be consistent. Our approach relies on the following observations and assumptions:
    \begin{enumerate}
        \item We only need to alter the object being reshaded and surroundings are left untouched: Note that this removes the possibility of inferring shadows for the object since they would likely fall outside the object's mask, however we decide to overlook this shortcoming since it is being researched in other works.
        \item The albedo of the item is invariant between its original context and its new context.
        \item The digital image is representable using only an albedo field and a shading field.
        \item The shading field is representable in a single channel (illumination is assumed to be uni-color)
    \end{enumerate}
\subsection{Cut And Paste}
    We next formalize the problem at hand in our formulation. Let $S$ be a source image containing some object in the region highlighted by a mask $M$. $M$ has the same shape as $S$ and holds a value of $1$ where the object exists and $0$ everywhere else. We can thus extract the object using element wise multiplication $X_o = M \odot S$. Now let $T$ be a target scene on which we wish to super-impose $X_o$ at some arbitrary location $L$ parametrized by horizontal and vertical offsets as well as potentially a size scaling factor. For the sake of simplicity and brevity, we assume that the offsets are 0 and the scaling is unity; the following derivations would be similar but with modified masks and locations computed using the parameters we neglect. We define the Cut and Paste $CP(A, B, C)$ operation as the following:
    \begin{equation}
        CP(A, B, M) = M \odot A + (1 - M) \odot B
    \end{equation}
    $CP(S, T, M)$ simply superimposes the object $X_o$ in the target scene without modifying any other pixel. Note that we will use this operation later for other fields and it follows the same principle albeit in fewer channels. The Cut And Paste operation allows us to focus on generating a reshading exclusively in the masked area, which is the motivation behind our first assumption.
\subsection{Albedo Invariance}
The second observation relates to the albedo $\rho_o$ of the object $X_o$ in question between its representations in the original context $S$ and the target context $T$. Let $X_o^*$ be the best possible representation of $X_o$ in $T$ in a way that is indistinguishable to the human eye (Think of this as the potential ground truth). We argue that the albedo $\rho_o^*$ of the optimal representation is exactly equal to the albedo of the initial object. We therefore should not require our network to learn information about the albedo since it is not required. The model would only output an intermediary field that is then used deterministically to build the image representation as dicussed in the next section.
\subsection{Image Formation}
Typical image formation models include three separate terms that describe the observed light incoming at a camera. The terms are: the direct diffuse term, the specular term, and an inter-reflections term. Most works drop the inter-reflection term since it is hard to model and is often not too substantial to affect the image. The specular term describes bright patches that are direct reflections of light sources off of surfaces. Robust methods exist to filter out specularities, however the opposite is not true; our method would ideally be able to model such specularities on the surface of the reshaded object $X_o$. In order to keep the complexity of the pipeline in check, we also drop this specularity term since we would also need to develop a consistency metric to judge the addition (or lack) of bright patches on the object. Lastly, there is the diffuse term which is tightly coupled to the albedo field over the object, along with the geometry of the object, and the incident light orientation. In fact, there exists a continuum of albedo fields and incident light fields over the object, parametrized by the light wavelength: the material reflects different amounts of light depending on the color of the incident illumination which in turn can be composed of different distributions of colors. Cameras collapse these continuums into quantized regions that are often centered around the human perception limits (R, G, and B fields). We can therefore model the diffuse term using 9 degrees of freedom according to the following:
\begin{equation}
    \begin{aligned} 
        I_R(p) &= \rho_R(p) * \left( N(p) . S_u \right) *S_R(p) \\
        I_G(p) &= \rho_G(p) * \left( N(p) . S_u \right) *S_G(p) \\
        I_B(p) &= \rho_B(p) * \left( N(p) . S_u \right) *S_B(p)
    \end{aligned}
\end{equation}
The $I(p)$ channels are what is observed in the image at pixel position $p$. $\rho(p)$ represents the aggregated albedo \footnote{The aggregation is a function of the camera sensitivity and sensor properties} for the quantized region (R, G or B). $N(p)$ is the normal at pixel location $p$ consisting of a 3-vector (or 4-vector in homogeneous coordinate frames). $S_u$ is a unit vector encoding the incident orientation of the light, parallel to the direction between the 3D point observed and the light source. Lastly the $S$ terms encode the color of the incident light after going through the described quantization. In our model and for simplicity of generating the feature extraction models, we assume that the incident light is uni-color, i.e $S_R = S_G = S_B$ which makes it representable in one channel. In fact, we reduce the image $I$ to be represented be the product of an albedo vector and a vector encoding both geometry and shading information. In other words:
\begin{equation}
    I(p) = \rho(p) * S(p)
\end{equation}
Or, in terms of the fields:
\begin{equation}
    I = \rho \odot S
\end{equation}

\section{RPNR}
Having described both our motivation and setup, we now describe the pipeline which is used to generate the reshading.\par

\subsection{Auxiliary Models}
\paragraph{Robust-Perception}
We desire a network that encodes observed objects into a feature space that is robust to changes in shading. In other words, the model should output highly similar encoding for two pictures of the same object but under different lighting conditions. To achieve such a model, we decide to fine tune a pre-trained classification model to make it internally consistent for a given object under different lighting. We describe this operation formally: Consider a classifier network $C$ with some internal layer $L_i$; also consider a set of images $X_{0..n}$ of the same scene but under different lighting conditions. Then we desire that:
\begin{equation}
    \begin{aligned}
        C(X_0) &= C(X_1) &= ... &= C(X_n) = Y\\
        L_i(X_0) &= L_i(X_1) &= ... &= L_i(X_n)
    \end{aligned}
\end{equation}
We fine tune AlexNet on a dataset containing diver object under various lighting changes.
\paragraph{Albedo-Shading Net}
We desire a network that extracts the albedo and shading fields from a given image. The model should simply output 4 fields (3 albedo, 1 shading) to closely model the observed image. This problem is very similar to a segmentation problem and most SOTA works employ a UNet to extract such features. We develop such a model and train it on syntetized data according to the same paradigms used in the work by Forsyth \cite{cutandpaste}: We generate Mondrian patches (rectangular patches of color), randomly rotate and scale the images, and multiply the resulting mosaic with a [0 - 1] perlin noise to emulate shading\footnote{This is valid since shading is generally slow variation}. We train the model to learn the inverse mapping (ie: extract $\rho$ and $S$ from $I$)
\paragraph{Normal-Shading Discriminator}
We desire a network that is capable of detecting inconsistent shading for some input, as well as the localization of that inconsistency. We choose to employ a U-Net based discriminator to detect bad shading, or a bad shading-normal correspondence. The discriminator would have as its input the output of the Shading Network, as well as the extracted normals using the same method used in \cite{cutandpaste} formulated by Nekrasov et al \cite{Nekrasov}. Given a normal field $N(p)$ and a shading field $S(p)$, the discriminator yields a measure of agreement between these two fields globally, as well as per-pixel.
\begin{equation}
    D(N, S) = \Bigl\{\ \begin{array}{ c c }
1 & \text{consistent}\\
0\leq x<1 & \text{inconsistent}
\end{array}
\end{equation}
\subsection{DIP}
The fundamental piece of the pipeline is the DIP that we train to generate the needed shading field. Contrary to previous work, and to avoid collapsing to trivial solutions, we stick to the proposition by Ulyanov et al \cite{DIP} to not provide the model with a prior different than the model parameters and architecture. The DIP is operating to solve a penalized inpainting problem since we provide it with a masked shading field, and require that the inpainting be consistent with the normals of the object being reshaded.
\subsection{Output Formation}
To generate the final output (combined images). We do the following: Let $S^*$ be the generated shading field from the DIP, then the final image formed is 
\begin{equation}
    Y = CP(M, \rho_o \odot S^*, T) = M \odot \left(\rho_o \odot S^* \right) + (1- M) \odot T
\end{equation}
Where $\rho_o$ is the infered albedo using the Albedo-Shading Net. This formulation restricts the problem immensely which is helpful us when trying to implement such models into AR systems directly. The albedo and shading of the output image are therefore as follows:
\begin{equation}
    \begin{aligned}
        \rho_y &= CP(M, \rho_o, \rho_T)\\
        S_y &= CP(M, S^*, S_T)
    \end{aligned}
\end{equation}
When running the pipeline on the above paradigm, one can observe that the generated shading field can sometimes introduce unwanted change to the content of the inserted fragment. We opt therefore to add a loss that curbs artifact creation: we choose perception based networks instead of deterministic features given the flexibility of the models.
\subsection{DIP Losses}
Having covered the pipeline, we can now showcase the training losses we use.
 Our required constraints on the output are now reduced to the following:
\begin{itemize}
    \item The generated Shading field must be consistent with itself. (i.e: the generated shading inside the mask must be consistent with the shading outside the mask).
    \item The generated Shading field must be consistent with the normal field inside the mask.
    \item The perceptual features of the object must not change after being reshaded
\end{itemize}
We design a loss for each one of these constraints and we formulate a DIP problem. We consider the masked target shading $S_x = (1-M) \odot S_T$ as a degraded version of the optimal shading field $S^*$. Furthermore, we consider the normal-shading consistency and the perceptual consistency as regularizers for the image. Although it suffices to use the information generated only (i.e: to run these consistencies inside the mask only), we opt to extract the features for all the generated image $Y$ as a means of simplicity and to isolate potential failure points. Recall that the feature extraction network is denoted as $f$, and the discriminator is denoted as $D$. Finally, the loss we use is as follows:
\begin{equation}
    \begin{aligned}
        \mathcal{L}_s &= \| \left(S_T - S_y\right)\odot (1-M)\|_2^2 \\
        \mathcal{L}_n &= -log\left(D\left(S_y, N_y\right)\right)\\
        \mathcal{L}_f &= \|f\left(Y\right) - f\left(C\right)\|_2^2\\
        \mathcal{L} &= \mathcal{L}_s + \mathcal{L}_n + \mathcal{L}_f
    \end{aligned}
\end{equation}
Where $N_y = CP(M, N_s,  N_t)$ is the extracted normals from the source object and the target scene cut and paste to superimpose them. $C = CP(M, S, T)$ is the naive cut and paste operation directly on the images.

\section{Progress and Experiments}
We trained the following networks according to the listed datasets and tasks:
\begin{enumerate}
    \item Shading Robust AlexNet: We fine-tune a pretrained AlexNet model to have consistent features for given scenes. We trained the model using a multi-task loss (consistency, and classification).
    \item We trained a discriminator to segment input shading fields in order to detect inconsistencies within the field. The data used is based on shading fields of landscapes that we distort using a perlin circle mask and perlin noise,
    \item Albedo-Shading network: we train a UNet that decomposes an input image into its Albedo and Shading components according to the model described previously. We use purely synthetic data based on Mondrian Patches (we use 10 orthogonal patches with random shape, location, and size) the patches are rotated, then the shading field is generated using Perlin Noise with a factor of 2 in each dimension. We construct the input of the model by combining these two fields, and we train the model to output the fields corresponding to these channels.
    \item DIP: we demonstrate using the DIP in an inpainting problem. Note that we enhance the original implementation in \cite{DIP} by batching over the noise generated, which makes our approach faster by at least 4 times.
\end{enumerate}
\paragraph{Results}
The pipeline is outputting valid shading fields however they are sometimes bland and lack the depth of the object (crevices, ruggedness, etc...).
\section{Discussion}
Our proposed pipeline isolates the required parts in a modular fashion. However, the choices of the model we choose for each loss/consistency affects the output greatly. The designed system and assumptions might be completely valid but the tools used to verify them might be flawed. It is indeed the case in our scenario: the lack of available datasets and computational resources limited our choices for modeling the losses. It has been shown that other approaches to the sub-problems yield better results, and given that the pipeline does indeed output believable shadings, but they are lackluster in quality. We claim that with better modeling of the losses, the output of the pipeline will necessarily improve. Additionally, the image formation model is limited in scope since it does not factor inter-reflections, nor the possibility of coloured light. Developing a system for inferring Albedo and COLORED shading would allow us to break away from our simplistic approach however no work has been done to the best of our knowledge to address this problem.

\end{multicols}
\newpage
\bibliographystyle{unsrt}
\bibliography{ref.bib}  
\end{document}